\documentclass{article}

\usepackage[preprint]{neurips_2019}

\usepackage[utf8]{inputenc} 
\usepackage[T1]{fontenc}    
\usepackage{hyperref}       
\usepackage{url}            
\usepackage{booktabs}       
\usepackage{amsfonts}       
\usepackage{nicefrac}       
\usepackage{microtype}      
\usepackage{graphicx,wrapfig}
\usepackage{caption}
\usepackage{subcaption}
\usepackage{amsmath,amssymb}

\title{Conditional Generation of Synthetic Geospatial Images from Pixel-level and Feature-level Inputs}

\author{%
    Xuerong Xiao\thanks{Equal contribution. SG is corresponding author. Alternative EMail: \texttt{swetava@cs.stanford.edu}} \\
    Stanford University\\
    \texttt{xuerong@stanford.edu} \\
    \And
    Swetava Ganguli\footnotemark[1], Vipul Pandey \\
    Apple \\
    \texttt{\{swetava,vipul\}@apple.com} \\
}

\begin{document}

\maketitle

\vspace{-1.0cm}
\section{Introduction}\label{section::introduction}
\vspace{-0.3cm}
Dearth of labeled data for training supervised deep learning models for real-world applications of computer vision plagues the large-scale deployment of machine learning models in many domains; including problems in geospatial analysis and remote sensing. As an example, detecting infrequent events or changes like road closures, road blocks, junctions changing to roundabouts, temporary turn restrictions, etc. are critical to keep a geospatial mapping service up-to-date in real-time and can significantly improve user experience and above all, user safety. Obtaining labels to train supervised models to detect infrequent mobility change events is expensive in time and money for a multitude of reasons \cite{xiao2020vae}. An inexpensive solution is synthetic generation of training data and labels based on user-provided conditional inputs that can be easily manipulated. In \cite{xiao2020vae}, we propose a novel deep conditional generative model that can synthetically generate various types of semantically rich, image-like representations of GPS trajectory data (e.g., CRM, HCRM discussed below) by conditioning simultaneously on pixel-level (e.g, road network) and feature-level (e.g., desired observation time interval) conditional inputs. Detection (e.g., pedestrian crosswalks, road centerlines) and classification (e.g, landcover, vegetation) of geospatial features are routinely cast as canonical tasks in computer vision such as object detection, semantic segmentation, instance segmentation, etc. Analogously, detecting locations affected by the aforementioned changes in mobility can be cast as a semantic segmentation task from image-like representations of privacy-preserving GPS trajectory datasets \cite{xiao2020vae}. Consider a raster representation of the earth's surface created with zoom-24 tiles \cite{z24tiledefinition}. Any contiguous set of $n \times n$ tiles can be considered as an $n \times n$ image whose pixels are the associated zoom-24 tiles. A \textit{count-based raster map} (CRM) is a single-channel image-like representation where the value of each pixel is the number of GPS trace occurrences in the zoom-24 tile corresponding to the pixel, counted over all trajectories during an observation time interval, $\tau = \Delta t$. If the count is bucketed based on the heading direction of the GPS traces into 12 buckets of $30^{\circ}$, and each bucket is represented as an individual channel, we obtain the 12-channel \textit{heading count-based raster map} (HCRM). Similar representations based on various attributes such as speed, modality of motion (e.g., walking, driving), etc. can also be computed. Specifically, we show that our model, called a VAE-Info-cGAN, can accurately generate CRM and HCRM for a fixed $\tau$ (feature-level input) just using the binary road network \cite{brndefinition} as the pixel-level condition. Furthermore, extending the methodology of \cite{shen2020interpreting}, we show that we can manipulate the latent representation to accurately generate samples for feature level inputs unseen in the training data. Specifically, we show that we can accurately generate CRM and HCRM for different values of $\tau$. 

\section{The VAE-Info-cGAN model proposed in \cite{xiao2020vae}}\label{section::vaeinfocganmodel}
\begin{wrapfigure}{r}{4.5cm}
        \centering
        \includegraphics[width=4.5cm]{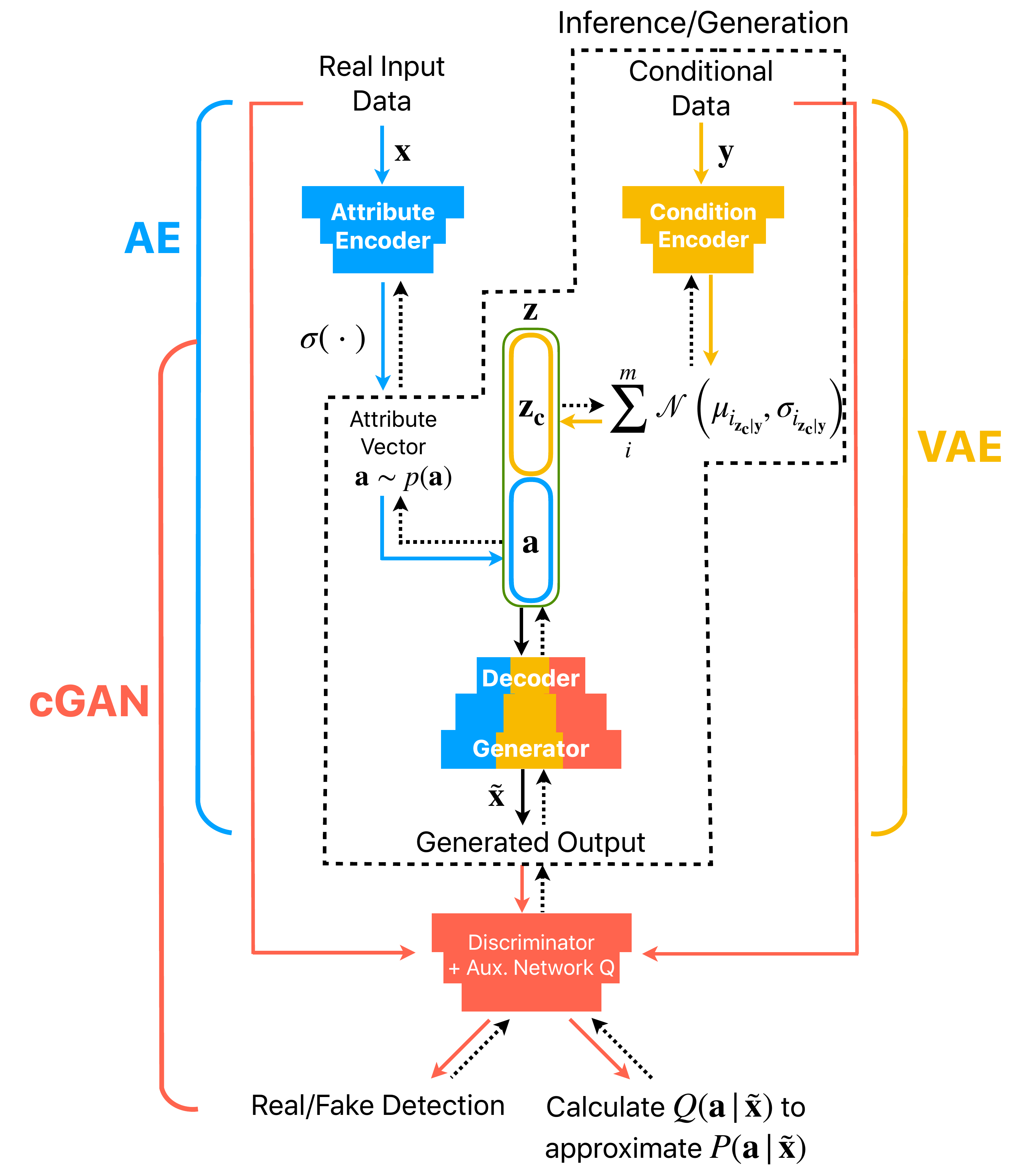}
        \caption{\centering{Schematic of the VAE-Info-cGAN model.}}
        \label{model_schematic}
\end{wrapfigure}

The goal of VAE-Info-cGAN is to learn to generate, $\boldsymbol{x} \in \mathbb{R}^{n \times n \times c}$ ($c=1$ for CRM and $c=12$ for HCRM), from the pixel-level condition, $\boldsymbol{y} \in \mathbb{R}^{n \times n \times u}$ ($u=1$ for binary road network), and feature-level condition vector, $\boldsymbol{a} \in \mathbb{R}^{d_a} \sim \mathbb{U}[0,1]$. As seen in the schematic of VAE-Info-cGAN (figure \ref{model_schematic}), the model combines a Variational Autoencoder (VAE) \cite{kingma2013auto} (yellow), a convolutional autoencoder (blue), and a conditional Information Maximizing Generative Adversarial Network (InfoGAN) \cite{chen2016infogan} (red) by sharing the the decoder of the autoencoder, the decoder of the VAE, and the generator of the InfoGAN. Once trained, only the components within the dotted area in figure \ref{model_schematic} are required for inference.
The model is trained using $(\boldsymbol{x},\boldsymbol{y})$-pairs collected from diverse geographical regions. During training, $\boldsymbol{a}$ is learnt directly from $\boldsymbol{x}$ by the autoencoder. The encoder of the VAE produces an embedding of $\boldsymbol{y}$, denoted by $\boldsymbol{z}_c$, which is modeled as a mixture of Gaussians. The concatenation of $\boldsymbol{a}$ and $\boldsymbol{z}_c$, denoted as $\boldsymbol{z}$, is fed to the generator of the InfoGAN to produce the output image, $\tilde{\boldsymbol{x}}$. The pixel values in CRM and HCRM are positive whole numbers which may result in a large dynamic range. Inputs, $\boldsymbol{x}$, are therefore log-normalized and the variational posterior, $p(\boldsymbol{x}|\boldsymbol{z})$, is modeled as a log-normal distribution. Let $L_{AE}=||\boldsymbol{x} - \tilde{\boldsymbol{x}}||_2^2$ denote the autoencoder's mean-squared loss, $L_{VAE}$ denote the evidence lower bound of the VAE component, $L_{gen}$ denote the non-saturating generator GAN loss, $L_{disc}$ denote the non-saturating discriminator GAN loss, and $L_{info}$ denote the information loss (as in \cite{chen2016infogan}). Define the total generator loss as $L_{G_{total}} = L_{AE} + L_{VAE} + L_{gen} + L_{info}$ and the total discriminator loss as $L_{D_{total}} = L_{AE} + L_{VAE} + L_{disc} + L_{info}$. The VAE-Info-cGAN is optimized by alternately minimizing the total discriminator loss and total generator loss.

\section{Qualitative and Quantitative Results}\label{section::results}
\begin{figure}[h]
    \centering
    \begin{subfigure}{0.28\textwidth}
        \centering
        \includegraphics[width=0.97\textwidth]{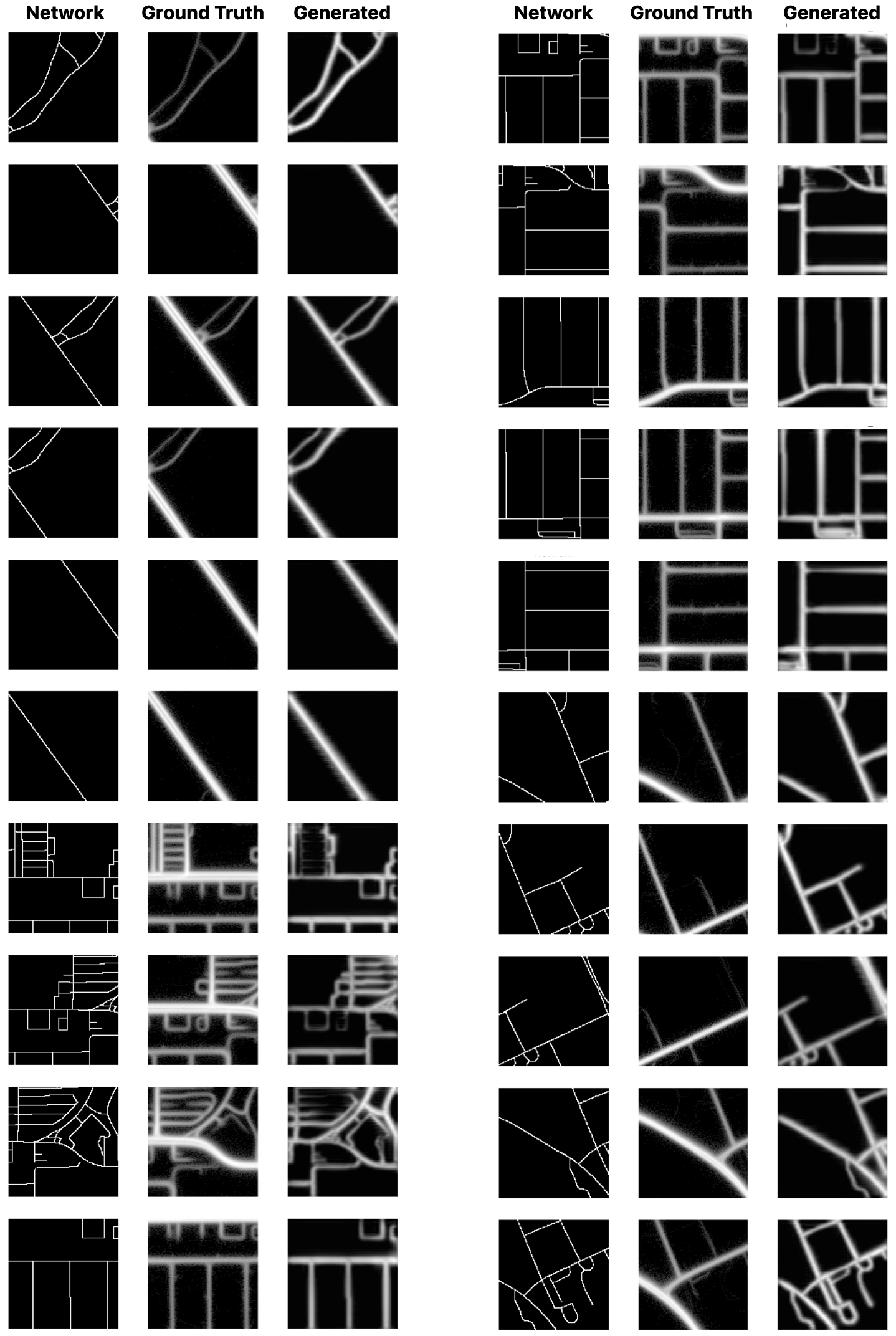}
        \caption{}
        \label{density_samples}
    \end{subfigure}
    \hfill
    \begin{subfigure}{0.26\textwidth}
        \centering
        \includegraphics[width=0.83\textwidth]{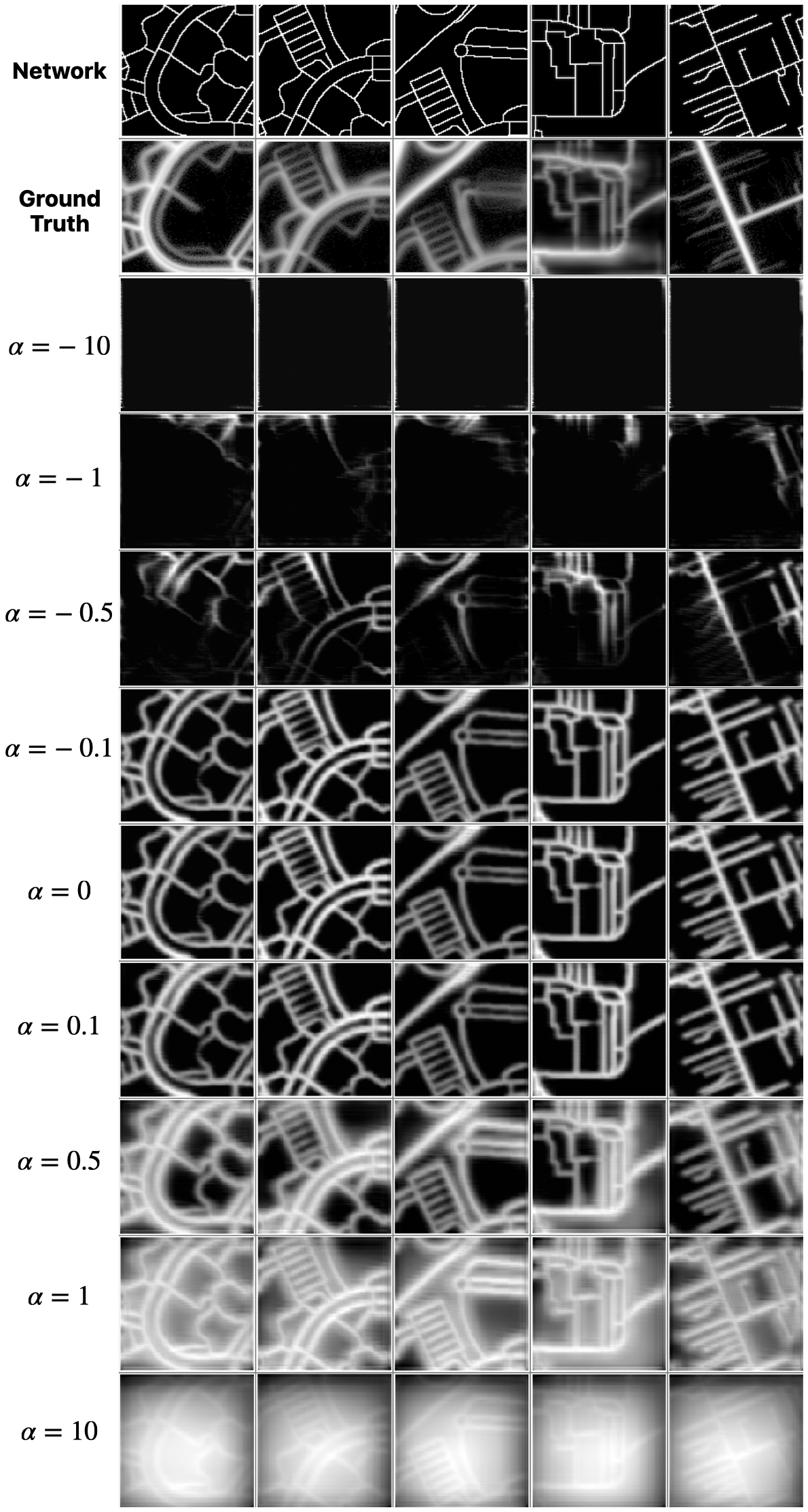}
        \caption{}
        \label{tuning_density_samples}
    \end{subfigure}
    \hfill
    \begin{subfigure}{0.44\textwidth}
        \centering
        \includegraphics[width=\textwidth]{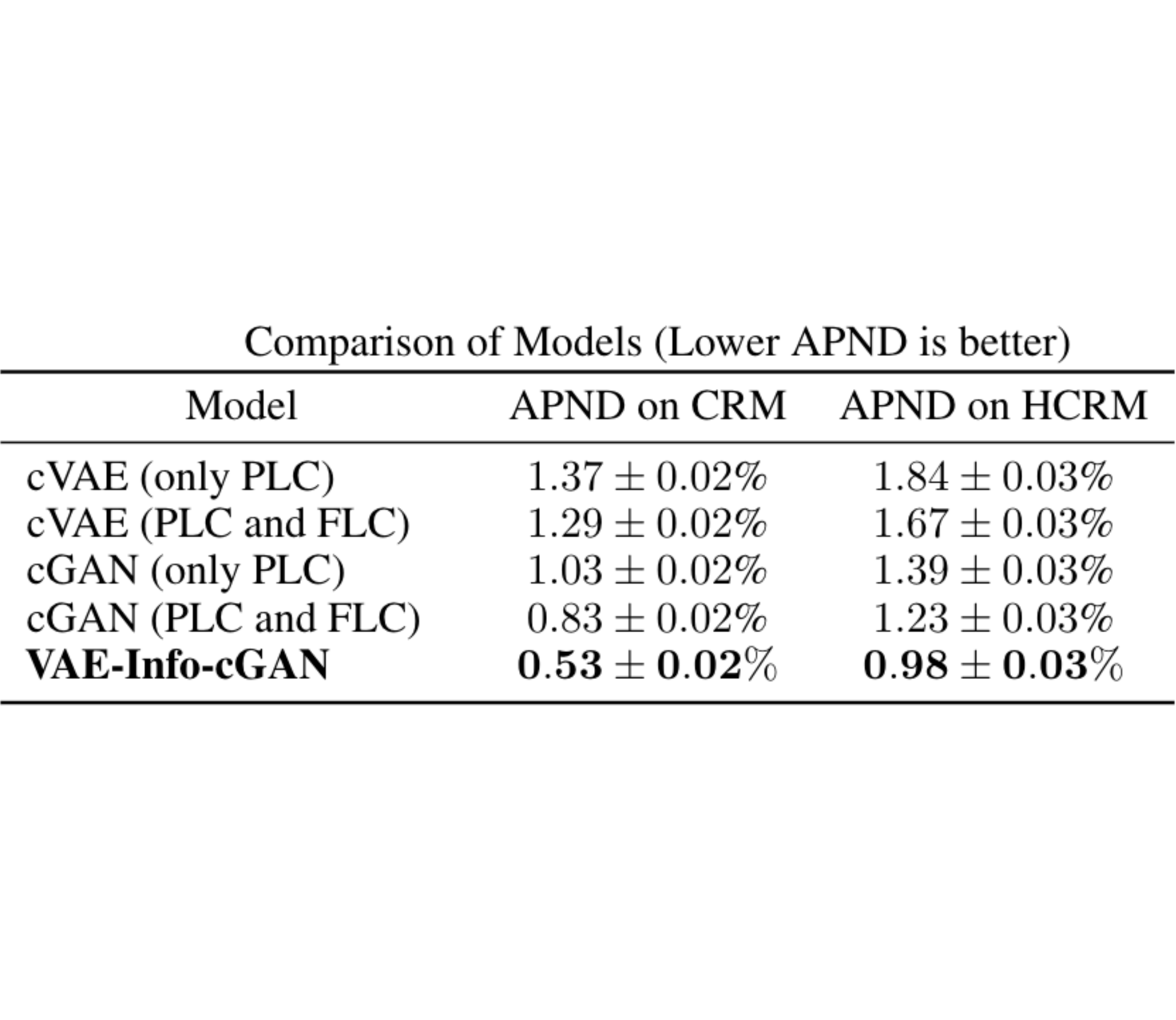}
        \caption{}
        \label{results_table}
    \end{subfigure}
    \caption{(a) Comparison of generated CRM with the ground truth. (b) Generated samples of CRM as $\alpha$ (a scalar proxy for $\tau$) is varied from -10 to 10. (c) Comparison of APND metric for generating CRM and HCRM between proposed and baseline models.}
    \label{results_figure}
\end{figure}

CRM and HCRM are computed in diverse geographical areas from \textit{probe data} \cite{probedata}. While generated samples for both CRM and HCRM \cite{generatingsamples} are available in \cite{xiao2020vae}, figure \ref{results_figure} shows qualitative results only for CRM. In all cases, the binary road network is the only pixel-level input. Figure \ref{density_samples} shows a comparison of ground truth and generated samples for 20 randomly chosen examples of road network (left column), ground truth (middle column),and generated CRM (right column). Figure \ref{tuning_density_samples} shows samples generated as $\tau$ (a feature-level input) is varied using a proxy scalar parameter $\alpha$. We found that when $\alpha=0$ corresponds to $\tau = \Delta t$, $\alpha=0.45$ corresponds to $\tau = 2\Delta t$. Evaluation of generative models must align with the application they were intended for \cite{theis2015note}. Since generating CRM and HCRM accurately is the goal and since the ground truth is available in the test set, we can quantitatively evaluate the accuracy of the generated samples using the average percentage normalized deviation (APND) metric defined as, $\text{APND} = \frac{1}{|\mathcal{D}_{\text{test}}|} \sum_{\boldsymbol{x} \in \mathcal{D}_{\text{test}}} \left[ \frac{\Vert \boldsymbol{x} - \boldsymbol{\tilde{x}}\Vert_2}{\Vert \boldsymbol{x} \Vert_2} \times 100 \right]$, where $\mathcal{D}_{\text{test}}$ is the number of examples in the test set. We compare the performance of the VAE-Info-cGAN to two variants of the conditional VAE (cVAE) and conditional GAN (cGAN): one with access to only the pixel-level condition (PLC) and the other with access to both the pixel-level and feature-level conditions (FLC). VAE-Info-cGAN outperforms its 4 other conditional generative model counterparts and may be viewed as the best of both the cVAE and cGAN models with additional modifications. 

\clearpage
\bibliographystyle{abbrv}
\bibliography{Paper_BayLearn2021}

\end{document}